\documentclass[letterpaper]{article} 
\usepackage{aaai25}  
\usepackage{times}  
\usepackage{helvet}  
\usepackage{courier}  
\usepackage[hyphens]{url}  
\usepackage{graphicx} 
\usepackage{natbib}  
\usepackage{caption} 
\usepackage{algorithm}
\usepackage{algorithmic}
\usepackage{booktabs}
\usepackage[table,xcdraw]{xcolor}
\usepackage{amsthm,amsmath,amssymb}
\usepackage{mathrsfs}
\usepackage{multicol, multirow}
\usepackage{graphicx}
\usepackage{bbding}
\theoremstyle{definition}
\newtheorem{definition} {Definition}
\usepackage{makecell}
\usepackage{newfloat}
\usepackage{listings}
\DeclareCaptionStyle{ruled}{labelfont=normalfont,labelsep=colon,strut=off} 
\floatstyle{ruled}
\newfloat{listing}{tb}{lst}{}
\floatname{listing}{Listing}
\title{VidEvent: A Large Dataset for Understanding Dynamic Evolution of Events in Videos}
\author{
    Baoyu Liang\textsuperscript{\rm 1, \rm 2},
    Qile Su\textsuperscript{\rm 1, \rm 2},
    Shoutai Zhu\textsuperscript{\rm 1, \rm 2},
    Yuchen Liang\textsuperscript{\rm 1, \rm 2},
    Chao Tong\textsuperscript{\rm 1, \rm 2} \thanks{Corresponding author}\\
}
\affiliations{
    \textsuperscript{\rm 1}School of Computer Science and Engineering, Beihang University, China\\
    \textsuperscript{\rm 2}State Key Laboratory of Virtual Reality Technology and Systems, Beihang University, China

    \{liangbaoyu96, zy2423342, shoutaizhu, liangyuchen, tongchao\}@buaa.edu.cn
}

\begin{document}

\maketitle

\begin{figure*}[h]
\includegraphics[width=0.99\textwidth]{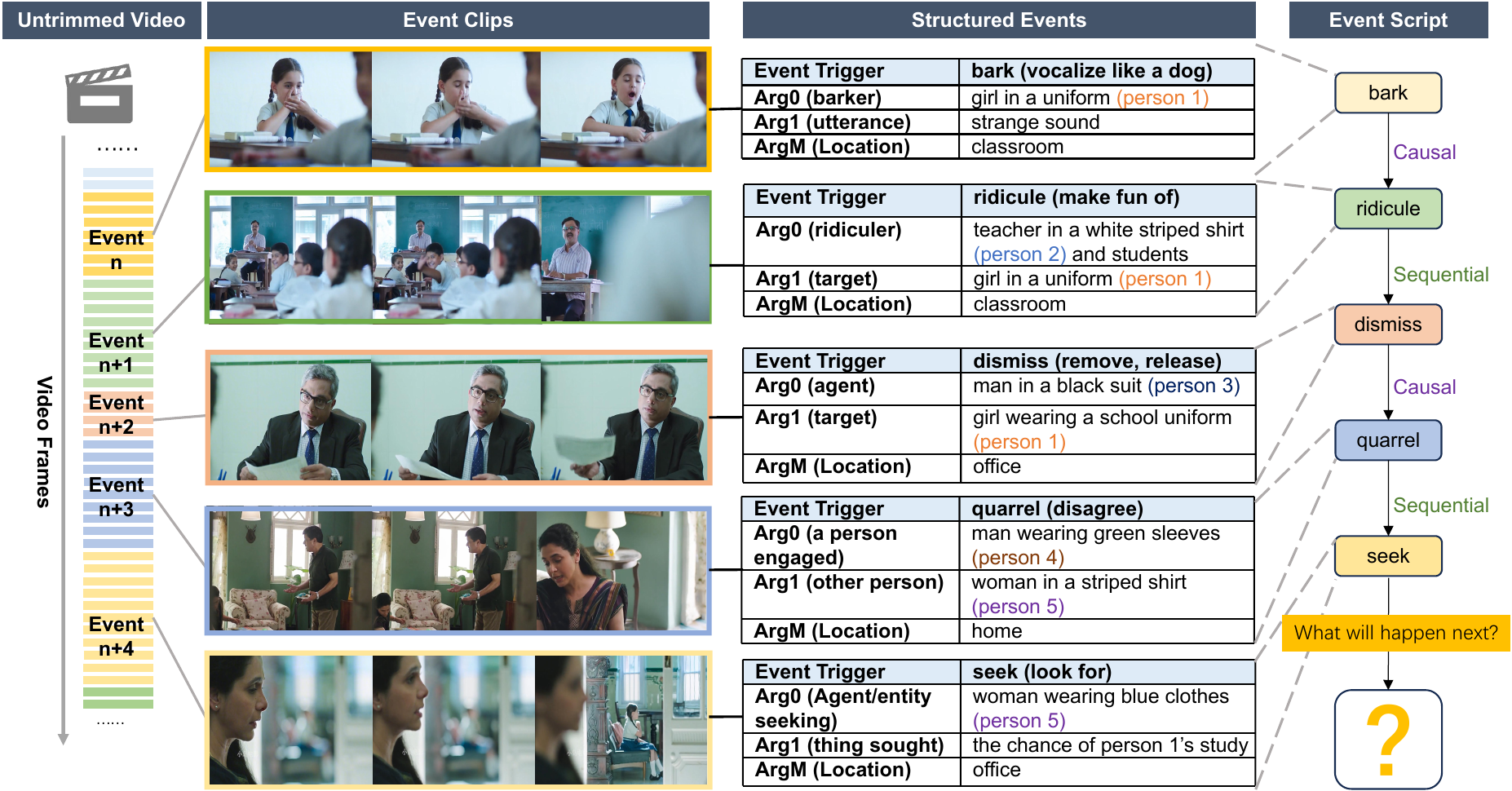}
\centering
\caption{The task of Video Event Understanding aims to extract event scripts with complete event structures and relations, and induct with event scripts. Here is an example in our proposed dataset VidEvent supporting the task. VidEvent provides untrimmed videos segmented with different event clips that vary in length. Each event clip is carefully labeled with a structured event including the event trigger like \textit{dismiss (remove, release)} and the event arguments like \textit{Arg0} and \textit{Arg1}. All the semantic roles participating in different events are co-referenced with a unique number like \textit{person 3} across the video. The relations among these events are also provided in VidEvent to support the construction and utilization of event scripts.}
\label{fig:dataset}
\end{figure*}
\begin{table*}
    \centering
    \resizebox{\textwidth}{!}{
        \begin{tabular}{llp{1.4cm}p{1.4cm}p{1.4cm}p{1.5cm}}
        \toprule
        \textbf{Task} & \textbf{Dataset}                                                                                    & \textbf{Event} & \textbf{Boundaries} & \textbf{Hierarchies} & \textbf{Evolution} \\
        \midrule
        AC            & Kinetics \shortcite{kinetics}, ActivityNet \shortcite{caba2015activitynet}, S-S \shortcite{goyal2017s2s}, HVU \shortcite{hvu2020}, Youtube8M \shortcite{youtube8M} & verb               & \XSolidBrush                        & \XSolidBrush                                      & \XSolidBrush                         \\
        ACL               & ActivityNet, Thumos \shortcite{Haroon2017THUMOS}, HACS \shortcite{Zhao2019HACS}, Charades \shortcite{Sigurdsson2016Charades}                           & verb               & \Checkmark                  & \XSolidBrush                                      & \XSolidBrush                         \\
        STD         & AVA \shortcite{Gu2018AVA}, EPIC-Kitchens \shortcite{Damen2018EPIC-Kitchens}, JHMDB \shortcite{Jhuang2013JHMDB}                                                & verb               & \Checkmark                  & \XSolidBrush                                      & \XSolidBrush                         \\
        VD                 & ActivityNet, Vatex \shortcite{Wang2020Vatex}, MSR-VTT \shortcite{xu2016msr} , LSMDC \shortcite{Rohrbach2016LSMDC}                                  & texts             &  optional  & \XSolidBrush                                      & \XSolidBrush                         \\
        VQA                      & MSRVTT-QA \shortcite{Xu2017MSRVTT-QA}, VideoQA \shortcite{Zhong2022VideoQA},  TVQA \shortcite{Lei2019TVQA}                         & texts     & \XSolidBrush                        & \XSolidBrush                                      & \XSolidBrush                         \\
        VR          & HowTo100M \shortcite{Miech2019HowTo100M}, DiDeMo \shortcite{Hendricks2017DiDeMo}, Charades-STA \shortcite{Gao2017Charades-STA}                                              & texts               & \XSolidBrush                        & \XSolidBrush                                      & \XSolidBrush                         \\
        VOG            & ActivityNet, VidSTG \shortcite{Zhang2020VidSTG},VID-sentence \shortcite{Chen2019VID-sentence}                                      & texts               & \XSolidBrush                        & \XSolidBrush                                      & \XSolidBrush                         \\
        VSRL                        & VidSitu  \shortcite{Sadhu2021VidSitu}                                                                                                               & structures           & \XSolidBrush                   &          \XSolidBrush         & pairwise                   \\
        VEU         &  \textbf{VidEvent  (Ours)}                                                & structures           & \Checkmark                  & \Checkmark             &  chain-like              \\
        \bottomrule
        \end{tabular}
    }
    \caption{A non-exhaustive summary on the video understanding tasks and datasets. }
    \label{tab:related-work}
\end{table*}
\begin{abstract}
Despite the significant impact of visual events on human cognition, understanding events in videos remains a challenging task for AI due to their complex structures, semantic hierarchies, and dynamic evolution. To address this, we propose the task of video event understanding that extracts event scripts and makes predictions with these scripts from videos.
To support this task, we introduce VidEvent, a large-scale dataset containing over 23,000 well-labeled events, featuring detailed event structures, broad hierarchies, and logical relations extracted from movie recap videos. The dataset was created through a meticulous annotation process, ensuring high-quality and reliable event data.
We also provide comprehensive baseline models offering detailed descriptions of their architecture and performance metrics. These models serve as benchmarks for future research, facilitating comparisons and improvements.
Our analysis of VidEvent and the baseline models highlights the dataset's potential to advance video event understanding and encourages the exploration of innovative algorithms and models. The dataset and related resources are publicly available at \url{www.videvent.top}.
\end{abstract}

%

\section{Introduction}

\label{sec:intro}

Events are at the center of human experience\cite{radvansky2017event}. We perceive events when observing, engage in events when acting, learn from events and use event knowledge to solve problems. While understanding events with our eyes is intuitive for humans, it remains a significant challenge for AI. Despite extensive research in natural language processing (NLP) on events, dealing with events in visual scenarios is still extremely difficult for AI models that are good at capturing static entities such as actions \cite{wang2023masked, duan2022revisiting, shi2023tridet} and moving objects \cite{wang2023visevent, wei2023autoregressive}, but lack understanding of dynamic events.

Unlike atomic actions or objects, events typically have the following three characteristics: (1) \textbf{Complex structure}. Events are composed of different constituents including participants, tools, time, and location, thus forming complex structures, which require a comprehensive understanding of the current situation. (2) \textbf{Various semantic hierarchies}. Events usually contain different semantic levels and relations. For example, 'A singer sings on stage' can be divided into several actions such as 'person standing on stage', 'person holding a microphone' and 'person making sound with mouth', while atomic actions can hardly describe such hierarchical relation. Distinguishing these actions from high-level events is difficult and requires understanding different semantic levels. (3) \textbf{Dynamic logical evolution}. Events are dynamic and evolve over time, following logical sequences. Understanding this evolution requires sophisticated event comprehension and commonsense reasoning, which current research finds challenging.

Recent video understanding methods have made progress in tasks like predicting actions, detecting objects, and describing visual scenes \cite{wang2023masked,wang2023visevent,wei2023autoregressive, duan2022revisiting,ko2023meltr}. However, few studies have focused on the analysis of structured and dynamic events, which are crucial to human cognition and could significantly advance computer vision (CV) research \cite{li2022clip}. Situation Recognition and Grounded Situation Recognition \cite{sadhu2021visual, khan2022grounded} are pioneering efforts in this direction, focusing on extracting event structures. Nevertheless, these tasks mainly emphasize event structures with limited consideration of the semantic hierarchies and dynamic logical evolution of events, which are crucial for cognitive understanding. We summarize these tasks and datasets in Table \ref{tab:related-work} and further analyze them in the Supplementary Materials due to the page limits.

In this paper, we further explore the understanding of events from visual scenes and propose the task of video event understanding that extract and induct with scripts of hierarchical, and dynamically evolutionary events and as is shown in Figure \ref{fig:dataset}. Video event understanding focuses on the logical evolution of events and thus encourages the extraction of highly conclusive events with high semantic levels other than the atomic actions and the prediction of the key relations that form complete logical chains. 

To support the task, we publish a new large-scale video event understanding dataset called VidEvent containing over 23,000 events and more than 17,000 relations from 1,110 movie recaps videos. VidEvent is characterized with complete event structures, macro event hierarchies and sophisticated event evolutionary chains. Events are ensured to have common complete structures and broad semantic hierarchies by a meticulous annotation process under the well-established Propbank framework \cite{kingsbury2003propbank} from highly concluded recap videos that are considered to have more complete structures, macro semantic levels, and more precise relations, compared with natural video or films. Relations among events are also carefully annotated following a strict labeling standard to form long event chains that ensures complete narrative logic.

We also provide evaluation metrics and baselines as benchmarks for future subsequent research. These benchmarks are designed with consideration of the challenges, i.e., the complex event structures, various semantic hierarchies, and dynamic logical evolution based on transformers. We also present various video understanding methods including SlowFast \cite{feichtenhofer2019slowfast}, TimesFormer \cite{bertasius2021space}, CLIP \cite{radford2021clip}, ActionFormer \cite{zhang2022actionformer}, TriDet \cite{shi2023tridet} and LLMs and provide comprehensive results for this task.

 
Our contributions include: 

(1) We propose the task of video event understanding which includes four progressive subtasks, aiming to enhance the scene understanding capabilities in higher event level, bridge the gap of event understanding and inference between CV and NLP, and explore the leap of artificial intelligence from perception to cognition. We claim our work to be the first to support extracting highly-concluded events and analyzing long-term event evolution as far as we know.

(2) We release a large-scale dataset called VidEvent in support of the proposed task. Over 1,000 movie recaps are carefully annotated to extract over 23,000 events with higher semantic level and more than 17,000 relations with accurate evolutionary logic between events.  

(3) Baseline methods and evaluation metrics are presented, along with detailed comparative results with current video understanding methods to form a comprehensive benchmark for future research.

\section{Task: Video Event Understanding}

\label{sec: task}
Analyzing the dynamic complex events is still a challenging task in computer vision depict the extensive research in the area of NLP. In face of the complex structured format, various semantic hierarchy, and dynamic logical evolution of events, we propose our task of video event understanding, aiming to extract the well-structured dynamic events with different semantic levels and further analyze the relations among them to supply the logic inference over the historical development of events. 

\subsection{Formal Definition}

We provide the following definitions to describe the task: 

\begin{definition}[Event Clip] Given a video $V$ with $T$ frames, an event clip $C \subset V$ is defined as a trimmed video clip in $V$ that describes a certain event $E$ within the event boundaries of $\mathbf{b}=\{b, e\}$, where $b \geq 0$ and $e \leq T$ are the beginning and ending frames of $E$.
\end{definition}
\begin{definition}[Event Structure] An event can be represented with an event structure $E=(trg, ARG)$, where an event trigger $trg$ is a word or span expressing the event. The argument set $ARG=\{{arg}_0, {arg}_1, \dots, {arg}_k\}$ contains the arguments corresponding to $trg$ that describe the detailed elements of the event. We represent a set of $E$ as $\mathcal{E}$.
\end{definition}
\begin{definition}[Event Relation]
    $R=\{r_1, r_2, \dots, r_m\}$ is defined as a set of relations, where each relation $r_j$ is a binary relation $r_j: \mathcal{E} \times \mathcal{E} \rightarrow \{0, 1\}$. That is, $r_j(E_i, E_k)=1$ indicates that the relation $r_j$ holds between events $E_i$ and $E_k$.
\end{definition}
\begin{definition}[Event Script with Relations]
  An event script with relations $\mathcal{S}=(\mathcal{E}, R_\mathcal{S})$ can be seen as a structured representation of sequences of events that have relations in a particular context or scenario, where $\mathcal{E}$ is a set of events and $R_\mathcal{S} \subseteq R$ represents the set of relations that hold between the events in $\mathcal{E}$.
\end{definition}
\begin{definition}[Script Event]
    An event $E_i$ in one event script $\mathcal{S}$ is called a script event.
\end{definition}
\begin{definition}[Video Event Understanding]
    The task of video event understanding is to understand the event scripts $\mathcal{S}$ from $V$ with all the components of events $\mathcal{E}$ and relations $R_\mathcal{S}$, and make use of $\mathcal{S}$ to predict next unseen events $\hat{E}$. It requires a comprehensive solution to segmenting the video $V$ into event clips $C$, extracting the complete event structure $E$ from $C$, analyzing the relations $R$ to form event scripts $\mathcal{S}$ and make predictions with $\mathcal{S}$.
\end{definition}

Unlike event understanding in NLP that has clear event boundaries within sentences and explicitly presents event triggers, arguments and even relations using words, understanding events and scripts in videos poses unique challenges due to the nature of video in terms of complex structure, semantic hierarchies and dynamic evolution of events. We further discuss these challenges and our considerations in simplifying the assumptions when designing the task.

\subsection{Timescale of Event Clips}

Events possess rich semantic hierarchies, and the amount of information a video can convey per unit of time can vary significantly. In previous work, a fixed timescale of 1 or 2 seconds has often been used to identify atomic actions \cite{Gu2017AVAAV} or video situations \cite{sadhu2021visual}. However, a fixed timescale might inadvertently capture incidental atomic actions or fragment a cohesive event, thereby complicating event analysis and inference.Therefore, we advocate that event clips at different hierarchical levels should be represented with varying timescales. In addition, in order to better facilitate reasoning about the evolution of events, we tend to choose longer timescales for event clips that contains events of higher hierarchies, while preserving the independence of each event. This is achieved by incorporating specific criteria into the annotation guidelines and conducting rigorous annotation reviews (see Supplementary Materials for detailed annotation criteria and the pipeline). This poses challenges of segmenting the events at a holistic level rather than merely dividing event clips based on scene transitions, character changes, or atomic actions.

\subsection{Event Structures and Correference of Arguments}


An event is conceptualized as an event trigger paired with a set of corresponding arguments. The triggers are derived from PropBank \cite{kingsbury2003propbank}, a comprehensive semantic role labeling corpus for English verbs. PropBank offers a detailed lexicon of verbs, each associated with a range of possible argument roles. We adopt them as our event vocabulary and the corresponding roles as argument templates. In addition, co-reference of entities is critical for analyzing event evolution across an event script. Previous work \cite{Sadhu2021VidSitu} have relied on imposing strict constraints as enforcing identical textual representations for co-referential arguments. However, this can be impractical in the context of long-duration videos or extended event scripts, where salient features of one character may vary over time. Hence we introduce a unique identifier for each person-related entity to facilitate consistent co-reference throughout the video in order to preserve the integrity of event reasoning in long event scripts.


\subsection{Event Relations}

Event relations are defined based on temporal, causal, conditional, coreference, and subevent relations, following the latest event relation extraction dataset MAVEN-ERE \cite{wang2022maven}. We introduce an additional conditional relation to specifically capture cases where one event is a prerequisite for the occurrence of another event. If Event B appears after Event A in a video, Event A: (1) occurs before Event B temporally in a temporal relation; (2) directly leads to the occurence of Event B in a causal relation; (3) is the essential condition, but is not directly responsible for the occurrence of Event B; (4) refers to the same event with Event B in a coreference relation; (5) includes Event B in a subevent relation. We prioritize causal and conditional relations over temporal ones because they inherently involve temporal aspects and play a more significant role in understanding event evolution. Unlike in NLP, analyzing event relation with videos lacks explicit or implicit connectives and referential cues that hint at relations between events. Therefore, a comprehensive understanding and inference of event semantics and inter-event relations are essential.


\subsection{Script Event Induction in Vision}
Inducting script events origins from NLP, with the aim of inferring the next possible events with current known events. Visual data can provide extensive details for understanding and predicting events in this task. However, the absence of explicit event representations and relational descriptions poses significant challenges. Given that this area remains largely unexplored, we propose a foundational taxonomy and simplify the task by introducing a single-choice setting. Formally, given the event clips $\mathcal{C}=\{C_1, C_2, \cdots, C_i\}$, a script structure $\mathcal{S}=(\mathcal{E}, R_\mathcal{S})$, and candidate events $\mathcal{M}=\{M_{0}, M_{1},$ $ \cdots, M_{4}\}$, it predicts the next possible event 
\begin{equation}
    \hat{E} = \arg\max_{E \in \mathcal{M}}(\text{Pr}(E| \mathcal{C}, \mathcal{S})). 
\end{equation}

We provide 5 candidate events in $\mathcal{M}$, two events are selected from a unified candidate pool, one event outside the event chain within the same video, and one event serves as a distractor by randomly replacing an event argument.


\section{Dataset: VidEvent}
To establish the database of Video Event Understanding task, we introduce VidEvent, a video event dataset composed of massive recap videos with structured annotations including diverse events, rich event arguments and precise event relations. The proposed dataset meets the demand for complex structures, various semantic hierarchy, and dynamic logical evolutionary. We further describe the detailed data and data analysis.

\begin{figure*}
    \centering
    \includegraphics[width=0.8\textwidth]{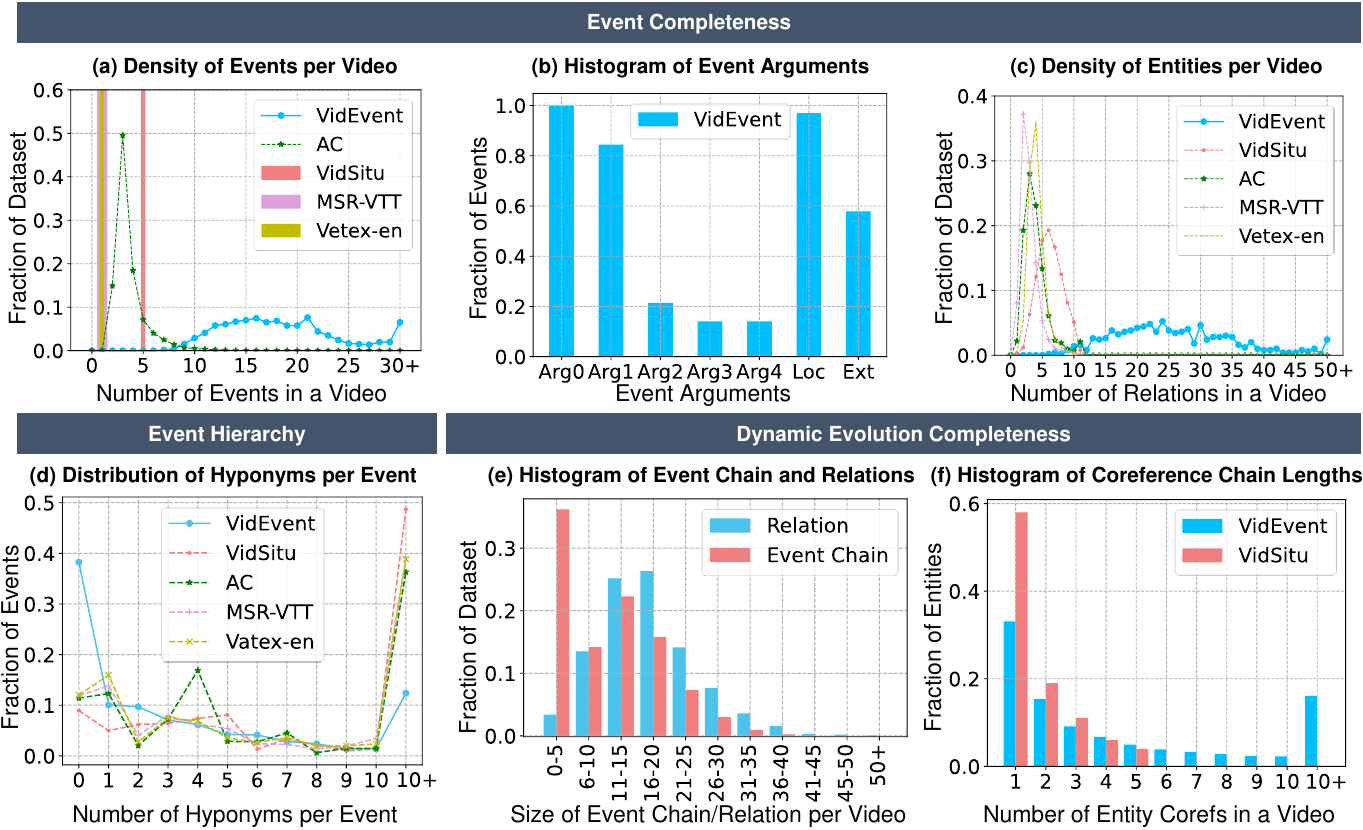}
    \caption{Data analysis. We emphasize on the \textbf{event completeness} of event structures (a-c), \textbf{event hierarchy (d)} and \textbf{relation completeness} of event relations and coreference chain lengths (e and f). AC represents ActivityNet Captions \cite{krishna2017dense}. }
    \label{fig:statistic}
\end{figure*}

\subsection{Data Composition}

\textbf{Movie Recap Videos.} VidEvent boasts a collection of movie recap videos. These videos succinctly summarize and condense the plots of original movies into several minutes. In contrast to the natural movie, recap videos offer highly concise summaries, with events and scenes aligning more clearly, and the logical progression of events being more pronounced \cite{singh2024recaps}. Typically, each sentence of the subtitles in the movie recap videos corresponds to one event, although there are cases where a single event may span several subtitles. In order to avoid the potential violation of copyrights, VidEvent only provides publicly accessible URL addresses for these videos, rather than the video files themselves.

\textbf{Annotations.} Each video in VidEvent is labelled with a JSON-formatted annotation. The annotations primarily consist of three types of information: 

(1) Video information. Video information containing video ID and URL is provided in this field. 

(2) Event clips. We include all annotated event in one video into event clips. Each event clip is annotated with one clip ID, the starting and ending time of the event, and the corresponding event structure. The entire event structure are sourced from Propbank. In addition, we add two extra arguments ArgM-LOC and ArgM-EXT that indicates the location and extent of the event's occurrence for each event to distinguish the details in visual events. All arguments are described in free texts with their salient features in the video. Each character is assigned with a unique identifier across the video, facilitating the comprehension of relations between events and characters. 

(3) Relations between event clips. Each relation within a video is identified by a relation ID and includes information including the relation type, the clip ID of the head event, and the clip ID of the tail event. The head event is defined as either a triggering event or a preceding event, while the tail event is considered as an affected, included or a subsequent event.

We provide more annotation examples as well as the data collection and annotation pipeline in the Supplementary Materials.

\subsection{Data Statistics and Analysis}

As shown in the Table \ref{tab:data_statistics}, VidEvent has a large number of videos and events with rich annotations and long event chains. A high annotation coverage rate in VidEvent indicates the high usability and coherence of the events. The average event chain length and coreference chain length suggest the presence of longer sequences of associations between events and complex character relations. All of these features support and challenge the long-term induction with visual events, which has not been address by current datasets.

\begin{table}[]
    \centering
    \begin{tabular}{lr}
    \toprule
    Data Statistics & Value \\
    \midrule
    Number of Videos & 1,110 \\
    Average Video Length & 1 min 22 s \\
    Average Event Length & 4.5 s \\
    Annotation Coverage of Events & 96\% \\
    Number of Events & 23,989 \\
    Number of Arguments & 80,822 \\
    Number of Relations & 17,525 \\
    Average Event Chain Length & 10.57 \\
    Average Coreference Chain Length & 4.67 \\
    \bottomrule
    \end{tabular}%
     \caption{Basic Data Statistics of VidEvent}
    \label{tab:data_statistics}%
\end{table}%

VidEvent is a dataset that contains events of complex structures, rich hierarchy, and dynamic logical evolutionary. To validate these characteristics, we compare VidEvent with other videos datasets containing text descriptions, MSR-VTT \cite{xu2016msr}, ActivityNet Captions \cite{krishna2017dense}, Vatex-en \cite{wang2019vatex} and VidSitu \cite{sadhu2021visual}. Among them, VidSitu provides complete event structures, while the annotations in the other datasets are sentences or captions describing the video scenarios.

\textbf{Event structure completeness}. The factor of event structure completeness indicates whether the structures of events in the dataset are sufficiently comprehensive, containing adequate and diverse event triggers and arguments. We examine the density of events per video, as illustrated in Figure \ref{fig:statistic} (a), and the distribution of event arguments, depicted in Figure \ref{fig:statistic} (b). Compared to existing datasets, VidEvent tends to have significantly more events per video, and the distribution of its labeled event arguments closely aligns with those in Propbank, indicating the completeness of events and arguments in terms of quantity. Additionally, we investigate the density of different entities per video, as shown in Figure \ref{fig:statistic} (c). Even within a single video, VidEvent tends to feature a wider range of diverse entities, reflecting a richer and more diverse set of events and scenarios covered in the dataset. This suggests the completeness of events in terms of diversity.

\textbf{Event hierarchical diversity}. We introduce event hierarchical diversity to assess the breadth of semantic hierarchies encompassed by the events within the dataset as illustrated in Figure \ref{fig:statistic} (d). This is achieved by comparing the distribution of hyponyms associated with each event, as events characterized by a greater abundance of hyponyms, such as 'speak' and 'walk,' typically denote broader semantic scopes, thereby exhibiting a lower semantic hierarchy. The figure reveals that events within the VidEvent dataset demonstrate a sparse distribution of hyponyms, yet maintain a diverse range of semantic hierarchies. This observation contrasts with other datasets primarily focused on atomic actions.

\textbf{Dynamic evolution completeness}. VidEvent provides the annotations for the dynamic evolution of events through the logic relations among different events. To evaluate whether the logical developments of events are complete, we analyze the distribution of logical relations in VidEvent and the length of event chains as shown in Figure \ref{fig:statistic} (e). Compared with other datasets, VidEvent tends to have more extensive distribution of relations in terms of both amount and variety. We also perform an analysis on the length of coreference chain of semantic roles in VidEvent to identify whether the key characters are labeled out with the development of events. From Figure \ref{fig:statistic} (e), VidEvent has considerably longer coreference chains, indicating its completeness in event evolution. 

\begin{figure}[h]
    \centering
    \includegraphics[width=0.45\textwidth]{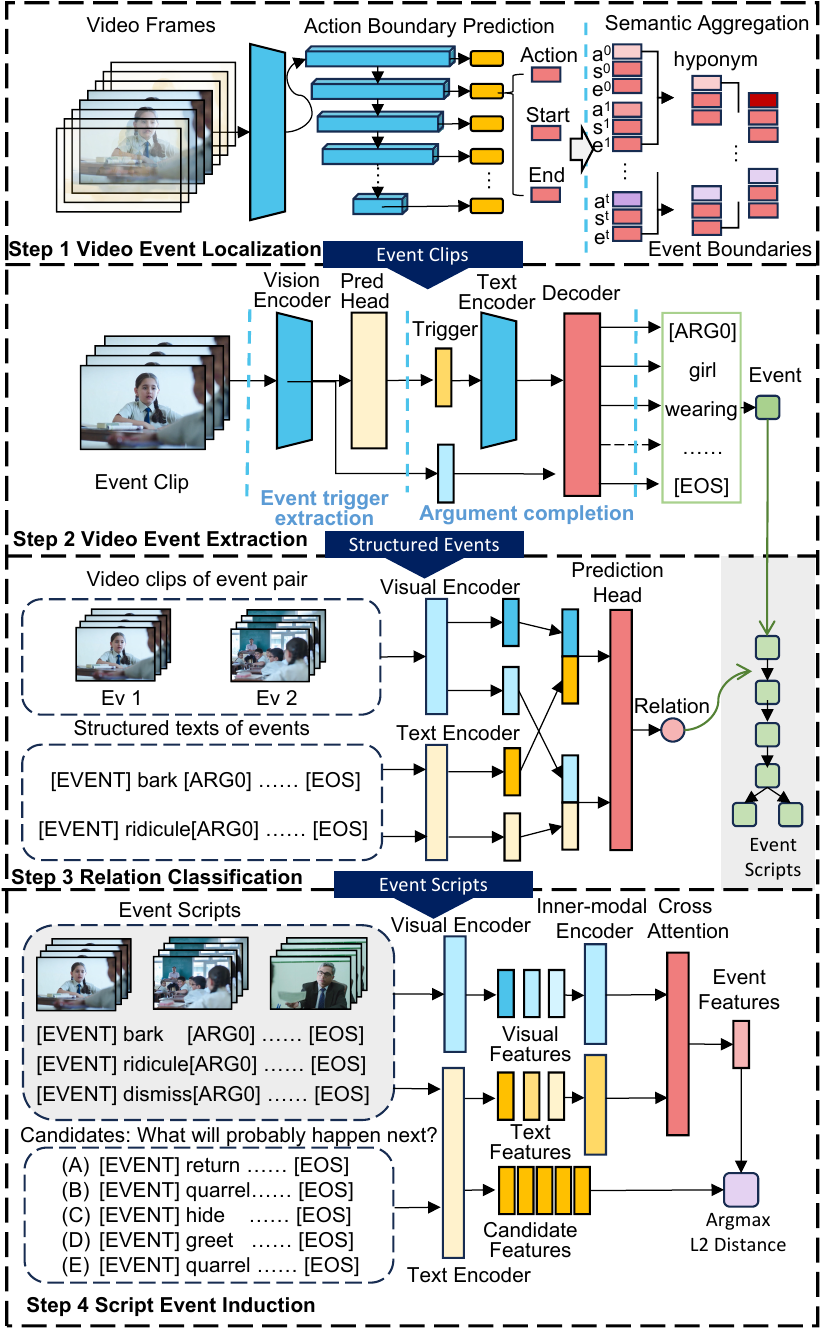}
    \caption{\textbf{The baseline models for the subtasks of video event extraction, video event localization, event ralation classification and event script induction.}}
    \label{fig:baseline}
\end{figure}

\section{Baselines}

The task of video event understanding encompasses the recognition and induction of event scripts. We design a fundamental framework, illustrated in Figure \ref{fig:baseline}, which consists of four essential steps that serve as a baseline for future comparative studies.


\textbf{STEP 1 (Video event localization)}. Given a video $V$, we require the model to recognize event boundaries $B=\{(s_i, e_i)\}$, where $s_i$ and $e_i$ are the start and end time of one event in order to obtain the event clips $C$. As baselines, we provide advanced action localization models such as TriDet \cite{shi2023tridet} and ActionFormer \cite{zhang2022actionformer} to extract atomic actions and their boundaries. All these models are trained from scratch with their original loss functions. To handle the semantic hierarchies of events, we design a semantic aggregation module that iterately aggregates boundaries of atomic actions with a semantic hyponym tree (see Figure \ref{fig:baseline}). We also provide an attention-based method for semantic aggregation based on TriDet. Details are provided in the Supplementary Materials. 

\textbf{STEP 2 (Video event extraction)}. 
For each event clip $C$, the model is expected to extract the complete event structure $E$ with an event trigger $trg$ and arguments $ARG$. We set $trg$ to be a verb from the vocabulary and $ARG$ a text sequence like 

\textit{[ARG0] girl wearing a uniform [ARG1] strange sound [ARGM-LOC] classroom [EOS]}.

We provide a step-wise method that initially predicts event triggers from individual video clips, followed by the synthesis of event arguments using both the video clips and the event trigger as shown in Figure \ref{fig:baseline}. We adapt widely-used video understanding models such as SlowFast, TimesFormer and CLIP\cite{feichtenhofer2019slowfast, bertasius2021space, li2022clip} as video encoders and transformer encoders like CLIP and RoBerta \cite{liu2019roberta} as text encoders. The encoded features are combined with a projection layer, and decoded with a transformer decoder like RoBerta. We also present a similarly-designed model in a joint prediction fashion that includes event triggers into the text sequences. 


\textbf{STEP 3 (Event relation classification)}. 
Given two extracted events $E_1$ and $E_2$ in their text form, and their event clips $C_1$ and $C_2$, we require the model to classify the event relation $r$ between them. 
To address this multimodal task, we propose a transformer-based baseline that incorporates both a visual encoder and a text encoder. The visual and text encoders can be selected from previous steps, such as SlowFast, TimesFormer, CLIP, RoBERTa, or other suitable models. Once the visual features $V_1, V_2$ and text features $T_1, T_2$ corresponding to events are extracted, these features are concatenated by event pairs. The concatenated features are then passed through a prediction head of two fully-connected layers, which is designed to predict the relation $r$ between the two events.

\textbf{STEP 4 (Script event induction)}. 
Given an event script $\mathcal{S}$ and corresponding event clips, we propose a two-branch architecture of vision and text as Figure \ref{fig:baseline}. Each branch contains an encoder for video or text to extract the corresponding features of $E_i \in \hat{E}$, followed by a transformer encoder to make inner interactions among events within one modal. The features extracted by these two branches are combined with a cross-attention layer to make inter-modal interactions and to generate the representation $\hat{F}$ for the predicted event. To represent the candidate events, we use share the text encoder to extract the representations $F^{M_j}$ of each candidate event $M_j$. Finally, we compute the L2 distances between $\hat{F}$ and each of the candidate event representations $F^{M_j}$, and choose the nearest candidate event $\hat{E}$ in distance as the answer. 

To bring the predicted future event representation closer to the correct candidate event while distancing it from the incorrect candidate event, we employ a triplet contrastive loss \cite{facenet2015schroff} to train the model. 

\section{Experiments}

Video event understanding supports evaluating in four steps: (1) video event localization, (2) video event extraction, (3) event relation classification, (4) script event induction with a given video.
\subsection{Metrics}
\textbf{Video event localization}. Video event localization is a task that combines boundary regression and type classification of events. For this task, we use mAP@IoU \cite{cheng2022tallformer, shi2023tridet, zhang2022actionformer, liu2022end} as the evaluation metric. We report the result at IoU threshold [0.1, 0.2, 0.3, 0.4, 0.5] and the average mAP computed at [0.1:0.5:0.1].

\textbf{Video event extraction}. In videos, the absence of explicit representations for event triggers and arguments in videos introduces language ambiguities, such as distinguishing between "chat" and "talk." To account for this ambiguity, we utilize Recall at 5 (R@5) and Precision at 5 (P@5) for event triggers, considering a prediction correct if any of the top 5 predictions match the ground truth. This approach helps to partially mitigate the impact of linguistic ambiguity. For evaluating the predicted arguments, we compute the average of METEOR \cite{Banerjee2005Meteor}, CIDEr \cite{Vedantam2015Cider}, and SPICE \cite{Anderson2016SPICE} scores for each event's arguments. These metrics are designed to account for the influence of synonyms and are robust against ambiguity.

\textbf{Event relation classification}. Event relation classification can be considered as a 6-way classification problem. Similar to that in NLP, we report top-1 accuracy, precision, recall and F1 score to evaluate the quality of predictions. 

\textbf{Script event induction}. Similarly to script event induction in NLP, we report the top-1 accuracy among five candidate events to evaluate the agreement between predictions and ground-truths. 
\subsection{Experimental Settings and Results}

We introduce the experiment results regarding to the baseline framework here. Implementation details, ablation studies and comparative results with other methods including LLMs are introduced in the Supplementary Materials. 
\label{exp}
\begin{table}[t]

\resizebox{0.48\textwidth}{!}{
\begin{tabular}{ccc|ccc|cccccccc}
\toprule
&   &   & \multicolumn{3}{c}{Event Trigger}  & \multicolumn{3}{c}{Event Arguments}\\

\multirow{-2}{*}{Type} & \multirow{-2}{*}{Encoder} & \multirow{-2}{*}{Decoder} & P@5& R@5  & F1@5 & METEOR& CIDEr & SPICE\\
\midrule
Step  & SF-RB    & VB    && {\color[HTML]{5B9BD5} }&& 0.24& 0.2 & 0.17\\
Step  & SF-RB    & RB  & \multirow{-2}{*}{0.51} & \multirow{-2}{*}{{\color[HTML]{5B9BD5} 0.46}} & \multirow{-2}{*}{0.48}& 0.31& 0.38&  {\color[HTML]{5B9BD5} 0.25} \\
\midrule
Step  & TF-RB   & VB    &&& {\color[HTML]{5B9BD5} }& 0.24& 0.21 & 0.17\\
Step  & TF-RB    & RB  & \multirow{-2}{*}{0.54} & \multirow{-2}{*}{0.44}& \multirow{-2}{*}{{\color[HTML]{5B9BD5} 0.48}} &  {\color[HTML]{5B9BD5} 0.32} & 0.3  & 0.24\\
\midrule
Step  & CLIP-CLIP&  RB & {\color[HTML]{5B9BD5} 0.62} & 0.3  & 0.41 & 0.31& {\color[HTML]{5B9BD5} 0.4} & 0.20\\
\midrule
Joint & SF& RB  & \textbackslash{}& \textbackslash{} & \textbackslash{} & 0.33& 0.39& 0.25\\
Joint & TF   & RB & \textbackslash{}& \textbackslash{} & \textbackslash{}  & {\color[HTML]{C00000} 0.33} & {\color[HTML]{C00000} 0.46} & {\color[HTML]{C00000} 0.25} \\
\bottomrule
\end{tabular}
}
\caption{Results of video event extraction. SF, TF, RB and VB are SlowFast, TimesFormer, RoBerta and VisualBert, respectively. The encoders of SF/TF-RB use SlowFast or TimesFormer as the visual encoder and RoBerta as the text encoder, while CLIP-CLIP means that the visual and text encoders are the visual and text parts of CLIP.}
\label{tab:res-event-classification}
\end{table}
\begin{table}[t]
\resizebox{0.48\textwidth}{!}{
\centering
\begin{tabular}{cccccccc}
\toprule
\multirow{2}{*}{Method}  & \multirow{2}{*}{Backbone}& \multicolumn{5}{c}{mAP@} & \multirow{2}{*}{mAP} \\
\cline{3-7}
 & & 0.1  &0.2  & 0.3  & 0.4  & 0.5 & \\
\midrule
ActionFormer & TSP  & 0.83& 0.81& 0.75& 0.59& 0.38 & 0.67\\
ActionFormer & SF & 0.76 & 0.73 & 0.63 & 0.50 & 0.30 & 0.59\\
TriDet  & TSP  & 0.81& 0.71 & 0.54& 0.34 & 0.17 & 0.51\\
TriDet  & SF    & 0.85 & 0.82 & 0.76 & 0.63 & 0.41 & 0.69\\
TriDet+Agg   & SF    & {\color[HTML]{C00000} 0.87} & {\color[HTML]{C00000} 0.85} & {\color[HTML]{C00000} 0.79} & {\color[HTML]{C00000} 0.70} & {\color[HTML]{C00000} 0.47} & {\color[HTML]{C00000} 0.74}\\
\bottomrule
\end{tabular}
}
\caption{Results of video event localization. mAP@$t$ is the average mAP at an IoU threshold $t$. mAP is the average mAP at the IoU threshold of [0.1:0.5:0.1].}
\label{tab:res-event-local}
\end{table}
\begin{table}[h]
\centering
\begin{tabular}{cccc}
\toprule

Method  & Text & Vision & Accuracy \\
\midrule
RoBerta &  \checkmark    &   & 0.55\\
TF    & &    \checkmark    & 0.23\\
SF& &   \checkmark& 0.25\\
TF+RoBerta &   \checkmark   &\checkmark   & {\color[HTML]{C00000} 0.56}\\
\bottomrule
\end{tabular} 
\caption{Results of script event  induction. SF and TF represents SlowFast and TimesFormer.}
\label{tab:res-script}
\end{table}
\begin{table}[ht]
\centering
\begin{tabular}{cccccc}
\toprule
Method & Text & Vision & P   & R   & F1  \\
\midrule
RoBerta&\checkmark & & 0.45& 0.37& 0.41\\
SF    & &\checkmark  & {\color[HTML]{C00000} 0.51} & 0.41 & 0.45\\
TF & &    \checkmark    & 0.41& 0.40 & 0.40\\
CLIP   &   \checkmark   & \checkmark& 0.46 & 0.38&  0.42 \\
SF+Roberta    &   \checkmark   &\checkmark & 0.50& {\color[HTML]{C00000} 0.46} & {\color[HTML]{C00000} 0.48} \\
\bottomrule
\end{tabular}
\caption{Results of event relation classification. SF and TF represents SlowFast and TimesFormer.}
\label{tab:res-relation}
\end{table}

\textbf{Video event localization}. We report mAP@0.1 to mAP@0.5 and the averaged mAP at [0.1:0.5:0.1] on the test set as shown in Table \ref{tab:res-event-local}. We observe that simply adding a semantic aggregation module increase the performance across all IOU thresholds. This aligns with our expectations regarding the semantic distinction between actions and events.

\textbf{Video event extraction.} We report P@5, R@5 and F1@5 on the test set for event trigger and METEOR, CIDEr, and SPICE for arguments. The results are shown in Table \ref{tab:res-event-classification}. The joint approaches outperform stepwise approaches sightly in all metrics, which may be attributed to the cumulative error inherent in the stepwise methodology. However, it is notable that joint approaches can hardly evaluate the performance in predicting individual event triggers, which can limit their potential applications to tasks that emphasize event triggers.

\textbf{Event relation classification}. We report the macro precision, recall and F1 score on test sets in Table \ref{tab:res-relation}.  The results indicate that both visual and textual modalities demonstrate comparable performance in event relation prediction, highlighting the potential effects of the visual modality in analyzing event relations. The results of SF+Roberta corroborate this conclusion and elucidate the complementary nature of the textual and visual modalities with the highest F1 score of 0.48 among these models.


\textbf{Script event induction}. 
In our experiments, we set the length of the event chains to 3 and train with a triplet loss. 
Other than the baseline model using two branches, we also present the results of simplified methods on one text or vision modal by removing one branch and the cross-attention layer in the baseline architecture in Table \ref{tab:res-script}. The results indicate a significant performance disparity between the textual and visual modalities in this task mainly attributed to the explicit expression of logical relations within the vast corpus used for pretraining textual models. Although multimodal learning shows only a slight improvement in performance, it also underscores the importance of improving the reasoning capabilities of the visual modality. 

 In summary, baselines show promise on the task of video event understanding. However, it is obvious that this task poses new challenges with a huge room for improvement.

\section{Acknowledgments}
This study is partially supported by National Natural Science Foundation of China (62176016, 72274127),
Guizhou Province Science and Technology Project: Research on Q\&A Interactive Virtual Digital People for Intelligent Medical Treatment in Information Innovation Environment (supported by Qiankehe[2024] General 058),
Capital Health Development Research Project(2022-2-2013), Haidian innovation and translation program from Peking University Third Hospital  (HDCXZHKC2023203), 
and Project: Research on the Decision Support System for Urban and Park Carbon Emissions Empowered by Digital Technology - A Special Study on the Monitoring and Identification of Heavy Truck Beidou Carbon Emission Reductions.

\bibliography{aaai25}

\end{document}